\documentclass[10pt, a4paper]{article}

\usepackage{svg}
\usepackage{makecell}
\usepackage[]{mdframed}
\usepackage[final]{lrec2026}
\usepackage{float}
\usepackage{booktabs}
\usepackage{tabularx}
\usepackage{subcaption}

\title{Creation of the Estonian Subjectivity Dataset: Assessing the Degree of Subjectivity on a Scale}

\name{Karl Gustav Gailit$^{1}$, Kadri Muischnek$^{1,2}$, Kairit Sirts$^{2}$} 

\address{University of Tartu \\
         $^{1}$Institute of Estonian and General Linguistics \\
         $^{2}$Institute of Computer Science \\
         \{karl.gustav.gailit, kadri.muischnek, kairit.sirts\}@ut.ee\\}

\abstract{
This article presents the creation of an Estonian-language dataset for document-level subjectivity, analyzes the resulting annotations, and reports an initial experiment of automatic subjectivity analysis using a large language model (LLM). The dataset comprises 1,000 documents---300 journalistic articles and 700 randomly selected web texts---each rated for subjectivity on a continuous scale from 0 (fully objective) to 100 (fully subjective) by four annotators.
As the inter-annotator correlations were moderate, with some texts receiving scores at the opposite ends of the scale, a subset of texts with the most divergent scores was re-annotated, with the inter-annotator correlation improving. In addition to human annotations, the dataset includes scores generated by GPT-5 as an experiment on annotation automation. 
These scores were similar to human annotators, however, several differences emerged, suggesting that while LLM-based automatic subjectivity scoring is feasible, 
it is not an interchangeable alternative to human annotation, and its suitability depends on the intended application.
 \\ \newline \Keywords{subjectivity, Estonian, annotation, LLM} }

\begin{document}

\maketitleabstract

\section{Introduction}

Understanding subjectivity is central to many Natural Language Processing (NLP) tasks that deal with human opinions, attitudes, and perspectives, such as sentiment analysis, stance detection, or opinion mining \citep{zhang2024sentiment}. Subjectivity analysis itself focuses on distinguishing subjective language, which conveys the author’s thoughts or feelings, from objective language, which presents facts and describes reality \citep{wiebe1994tracking}. Accurate identification of subjectivity thus contributes to the broader goal of modeling how people express evaluative or emotional meaning in text.

The most common approach to subjectivity detection is sentence-level analysis, as in the popular subjectivity dataset SUBJ \citep[]{pang2004}, which consists of sentences from movie reviews (labeled as subjective) and plot summaries (labeled as objective). Subjectivity can also be analyzed at the document level, for instance, detecting whether newspaper articles are subjective (such as opinion pieces) or objective (such as news) \citep[]{deft}. 

Creating reliable datasets is essential for subjectivity analysis, and two main approaches are typically used. The first relies on human annotators to label texts as subjective or objective. The second infers subjectivity from the text source, for example, assuming that movie reviews are subjective while plot summaries are objective. However, this latter method often produces noisy and unreliable labels \citep[]{czech}.

Most available datasets label subjectivity dichotomously---classifying texts as either objective or subjective. Yet human perception of subjectivity is gradual, not categorical, lying on a continuum between complete objectivity and complete subjectivity. Some previous work has tried to capture this variation using Likert-type scales \citep[]{wiebe2000, escouflaire}, but these still restrict the range of possible judgments. A continuous numeric scale could offer a more nuanced representation, allowing annotators to express subtle differences in perceived subjectivity.

This paper introduces an approach to \emph{document-level subjectivity annotation in Estonian}, a language for which no subjectivity dataset has previously been available. This approach is based on a theoretical approach previously suggested in \citet{gailit_towards_2025}. We describe the dataset creation, where annotators rated texts on a scale of 0--100 (fully objective to fully subjective), discuss the resulting annotation patterns, and present an initial experiment using large language models (LLMs) to explore automated subjectivity estimation.

\section{Creation of the Estonian Subjectivity Dataset}

This section gives an overview of the creation of the Estonian Subjectivity Dataset\footnote{\url{https://huggingface.co/datasets/tartuNLP/Estonian_Subjectivity}}, and the included data is described in Appendix C.
We describe the initial pilot project, the final selection of source texts, the main considerations behind the annotation procedure, and the annotation process itself. Finally, we present the annotation results through an analysis of inter-annotator agreements.

\subsection{Dataset Creation Principles}
Our goal was to collect subjectivity annotations that enable discriminating between subjective and objective texts in varying degrees. Therefore, instead of labeling texts categorically as subjective or objective, we set out to rate the degree of subjectivity on a scale where one end corresponds to completely objective texts and the other end to completely subjective texts. Although Likert scales are often used to annotate subjective tasks (such as sentiment), we considered the 5- or 7-point scales too restrictive and opted for a continuous scale ranging from 0 to 100, where low values close to 0 should be assigned to objective texts and high values close to 100 should characterize very subjective texts.
Using a sliding scale would also theoretically allow annotators to score more intuitively than having to choose from a small number of labels.

The aim was to create a dataset of 1000 texts annotated with subjectivity ratings. Prior to the dataset creation, a small-scale pilot study was conducted to test the source selection principles and the validity of the annotation method. 

\subsection{Pilot Project}

For the pilot project, a small dataset of 60 texts was selected from the Estonian National Corpus \citep{ENC}. Forty texts were sampled randomly---first by choosing a random URL and then a random text from that URL---with text lengths between 100 and 6,000 characters. In addition, 20 journalistic texts comprising 10 news stories and 10 opinion pieces were selected. Both news and opinion pieces came from two sources: Estonian Public Broadcasting (ERR), which is generally regarded as a neutral and objective source, and Uued Uudised (UU; New News), a right-wing alternative media outlet known for opinionated coverage. These sites were selected as representing opposite ends of the subjectivity-objectivity spectrum and both hosting news and opinion texts. We expected the random subset to have the full range of subjectivity, the news texts to be objective, and the opinion pieces to be more subjective, with ERR news being more objective than those from UU.

Two annotators, both female undergraduate linguistics students, were recruited to annotate the 60 texts. They were paid 100€ for this work. The annotation tasks were set up as a survey in LimeSurvey. The guidelines given to the annotators included a brief definition for both subjectivity and objectivity, an explanation on the scalar nature of subjectivity, the goal of the annotation task, and three example texts: one subjective, one objective, and one that would be between them on the scale. 

Based on a discussion with annotators after the project, the annotation task was deemed to be feasible, and the continuous rating scale was judged to be understandable and easy enough to use. The annotators expressed no issues with the texts themselves, including the maximal text length, though they noted that longer texts needed more focus and thought to annotate. This indicated that the full annotation project can use a similar upper limit to the text lengths as the longest texts used in this pilot project.

By inspecting the ratings given, the expected patterns were observed---the 40 random texts had varying scores, all ten opinion pieces were annotated more in the subjective end, and the five ERR news stories were annotated as more objective, while the five UU news stories were annotated with varying degrees of subjectivity, some as objective and some as subjective. With the high inter-annotator agreement as measured with the Pearson correlation coefficient of 0.86 (\textit{p}-value below 0.001), the small-scale annotation test was considered to be a success and showed that the method of using a numeric scale for annotating subjectivity is possible, allowing us to move forward to annotating the larger, 1000-document dataset.

\begin{table*}[t]
    \small
    \centering
    \begin{tabularx}{\textwidth}{llX}
        \toprule
         \bf Category & \bf Count & \bf Description \\
        \midrule
         News & 150 & Journalistic news articles from online media outlets (part of the 300 journalistic pieces).\\
         Opinions & 150 & Journalistic opinion pieces from online media outlets (part of the 300 journalistic pieces).\\
         Advertising & 331 & Texts that advertise a product, service, or provider.\\
         Social media & 130 & Social media posts, primarily forum messages and blog posts.\\
         Web journalism & 88 & Journalistic articles from the web (part of the 700 randomly selected web texts).\\
         Informative & 46 & Texts, where the goal is to provide factual information, such as encyclopedia entries and Wikipedia articles.\\
         Instruction & 43 & Texts, where the goal is to guide the reader through a process, such as recipes and tutorials.\\
         Review & 14 & Texts expressing user opinions and experiences with a specific product, piece of media, etc.\\
         Legal & 10 & Legal texts such as privacy policies and terms \& conditions.\\
         Literature & 8 & Literary texts, including fiction and poetry.\\
         Miscellaneous & 30 & Texts that do not fit into another category and that do not have enough representation to make a new category.\\
        \bottomrule
    \end{tabularx}
    \caption{Statistics of the categorical composition of the selected documents.}
    \label{tab:categories}
\end{table*}

\subsection{Documents Included}

For the full-scale dataset, a set of a total of 1000 texts sourced from the 2021 edition of the Estonian National Corpus \citep{ENC} was selected. All texts were constrained to between 100 and 6000 characters. Shorter texts were excluded, as they tended to not include enough information to provide meaningful subjectivity analyses, often consisting of a single sentence or even only the title of an article. Longer texts were excluded to make the annotation process simpler for the annotators, as, for instance, too long texts may cause instances where the annotator forgets the beginning of the text and bases their score only on the ending.

Of the 1000 texts, 700 were randomly selected from the full Estonian National Corpus 2021 \citep{ENC}. To ensure a wide variety of sources and topics, the texts were chosen by first selecting a random website represented in the corpus, and then a random text from that website. This led to 694 unique source URLs and a maximum of two texts per source URL.

The remaining 300 texts consist of journalistic texts published on the Internet, with 150 being news articles and 150 being opinion pieces. These were chosen from the Estonian National Corpus 2021 \citep{ENC} Feeds subcorpus and were manually checked to ensure relevance and suitability of the labels ``news'' and ``opinion piece''. As with the 700 Web texts, the journalistic texts were also chosen randomly. There are 106 unique source URLs for the news articles, with the most frequent URL being the source of 10 texts. For opinion pieces, the number of unique sources is 45, with the most frequent URL representing 24 texts. This reduction in the variety of source URLs was most likely caused by the low character maximum, as many of the opinion pieces in the corpus were longer than 6000 characters.

All texts were manually checked to ensure they were in Estonian and did not contain encoding errors or any issues that could cause the text to be hard to read. Additionally, private information of non-public individuals was pseudonymized. Phone numbers were all replaced with (+372) 51510105, a value visually similar to Estonian phone numbers, but not able to be used. Emails were replaced by the string "EMAIL", as several email addresses in the Estonian National Corpus were already anonymized using this method. Full names were replaced according to gender, with each person given a common Estonian name. This list of names is as follows: Juhan Mägi, Kaarel Märg, and Andrus Mesi for men, and Linda Mets, Anu Muru, Katrin Mingi, and Sandra Mäss for women.

When looking over the annotated set of texts, specific categories or genres began to stand out. These categories are not a part of the Estonian National Corpus and were assigned manually during checking, with each text getting a singular label that most dominantly describes the text.
It should be noted that this assignment of categories was only a secondary task and was only included since information on genre would be useful for analysis. Because of this, no previously established genre taxonomy was used.

As these texts were chosen randomly, the resulting dataset is imbalanced, with some categories being significantly overrepresented compared to others. The list of categories, together with their description and frequency in the dataset, is shown in Table~\ref{tab:categories}.

\subsection{Annotation Process}

Three annotators (individually referred to as A1, A2, and A3) were recruited to annotate the full dataset consisting of 1000 texts. Taking place in January 2025, the texts were sent in four batches of 250. The annotators were paid 250€ per batch, for a total of 1000€. Similar to the Pilot project, the annotation was done in the LimeSurvey environment, where the texts were presented in a random order. The annotators were asked to score each text using a sliding scale, where 0 marked an objective text and 100 a subjective text, and also select their certainty of this score, choosing between ``not at all certain'', ``partially certain'', and ``entirely certain''. The annotators were also given a field for optional comments, but explaining the thought process behind each annotation was discouraged to limit the influence of known causes of subjectivity, such as subjective adjectives, and therefore encourage a more holistic approach. The full instructions given to the annotators, as well as the translations to English, are provided in the appendix.

\subsection{Annotation Results}
\label{subsec:annotation_results}

\begin{figure}[!t]
\begin{center}
\includesvg[width=\columnwidth]{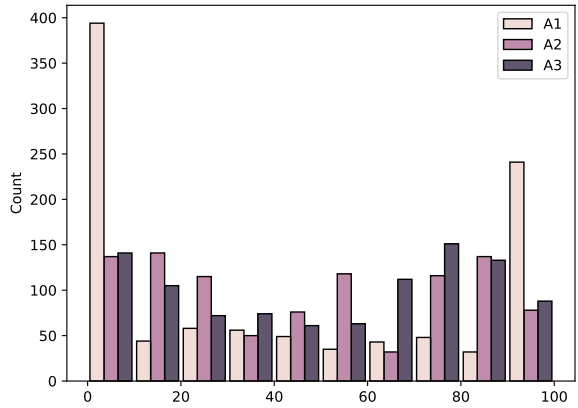}
\caption{Histogram of the subjectivity scores given by the three annotators}
\label{fig:score_hist}
\end{center}
\end{figure}

Between annotators, some degree of disagreement is expected, as subjectivity is a subjective topic---different people understand subjectivity in different ways, and their internal scales of subjectivity and objectivity vary accordingly. For instance, A1 was unique in preferring the ends of the scale, with 391 texts getting a score of 0 and 234 texts getting a score of 100. 
The histograms of the scores given by all three annotators are plotted in Figure~\ref{fig:score_hist}. 

\begin{figure}[!t]
\begin{center}
\includesvg[width=\columnwidth]{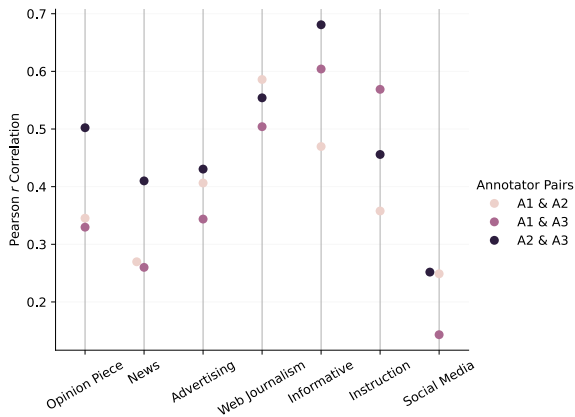}
\caption{Inter-annotator agreement between categories on the full dataset of 1000 texts.}
\label{cats1000}
\end{center}
\end{figure}

We assessed the inter-annotator agreements using the Pearson correlation coefficient. The Pearson correlations between the annotators were somewhat lower than observed in the pilot project: 0.570 (A1 vs. A2), 0.525 (A1 vs. A3), and 0.627 (A2 vs. A3). All of the correlations were statistically significant with \textit{p}-values all below 0.001.
Figure \ref{cats1000} showcases these correlations within categories that contained at least 30 samples. All categories have a large amount of variation in the amount of disagreement between the annotators. While the categories Web Journalism and Instructive have the highest average correlations, Social Media texts are the least correlated. The correlations in the lower range indicate that there might be texts that were judged qualitatively considerably differently by the annotators.

However, large differences between the subjectivity scores of a text indicate a need for further analysis. We first qualitatively examined the texts with the highest inter-annotator differences, but no clear patterns emerged.

Next, we examined the correlations between text length and the annotator subjectivity scores. Although the correlations were weak to moderate, they were all statistically significant (with \textit{p}-values all below 0.01). For the mean annotator score, the Pearson correlation with the text length in characters is 0.289, in words it is 0.342, and in sentences it is 0.367, meaning that the longer the text, the more likely it is considered to be more subjective than objective. The correlation with the average word length (text length in characters divided by the text length in words), on the other hand, is -0.514.

Additionally, we analyzed the correlations for each text category in the dataset. While several genres did end up with much lower correlations, the smallest being News, with the correlation of the length of the text in characters being only -0.005, others had much stronger correlations, with Legal texts having a correlation of 0.666 with the text length in characters. This shows that text length has a different impact on subjectivity annotations depending on the genre the text itself belongs to.

Finally, we looked at the confidence judgments collected during annotation, rated on a 3-point scale.
We hypothesized that texts with greater disagreement between annotators would also receive lower confidence ratings (that is, ``partially certain'' or ``not at all certain'').
However, the correlation between total annotator confidences and the sum of absolute score differences was weak (Pearson's = \emph{r} -0.09, \textit{p}-value = 0.125), suggesting that disagreement among annotators was largely independent of their confidence.
We also examined whether text length, measured in characters, words, or sentences, correlated with annotation differences. 
The correlations were all close to zero, indicating no systematic relationship, with \textit{p}-values above 0.646. Text length showed similarly weak correlations with total annotator confidence. These results suggest that text length was not a factor in the larger annotation differences.

\section{Secondary Annotation}
In order to further investigate possible reasons for large differences between annotator ratings, we aimed to re-annotate a subset of the dataset with especially high disagreement.

\subsection{Re-annotation}
\begin{figure}[!t]
\begin{center}
\includesvg[width=\columnwidth]{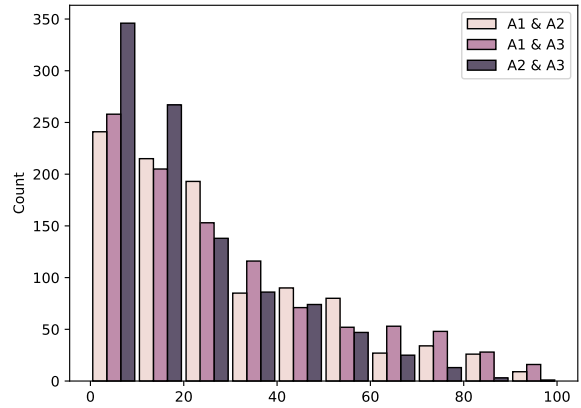}
\caption{Histogram of differences between each of the annotator pairs. Higher x-axis values indicated higher differences in annotators' scores. }
\label{fig:score_diff}
\end{center}
\end{figure}

Figure \ref{fig:score_diff} plots the histograms of score differences between all three annotator pairs. While most differences are in the lower range, there are a considerable number of texts where the scores are in different polarities of the scale (difference > 50). 

We selected for re-annotation texts where the difference between the scores of one annotator and the average of the other two annotators was larger than 50. This number was chosen since it indicates highly conflicting annotations, where the annotators had clearly preferred two different sides of the scale or, equally importantly, the edge of the scale against the middle of the scale. There were 220 such texts in total.

As a control, 30 additional texts were added. Half of the texts were selected to have a very low annotation difference (below 3), indicating similar annotations between all annotators, and there were 14 such texts in the dataset. Another half were selected to have a moderate but reasonable difference between the annotators (between 24 and 26), of which there were 16 texts. This amounted to the total re-annotation subset of 250 texts.

We invited the same three annotators to re-annotate this subset and obtained repeated annotations from two of them (A2 and A3), who were compensated 250€ each. The 250 texts were presented as a single batch, with instructions explaining that the goal was to examine how annotations might differ after four months. The annotators were not told that most texts had previously received highly divergent scores to minimize potential bias. After completing the task, each annotator participated in a short interview about their re-annotation process and the factors they felt influenced their scoring.

\begin{figure}[t]
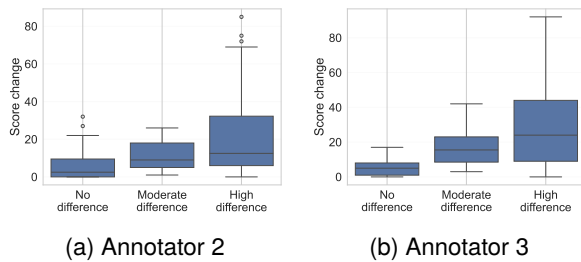

    \centering
    \begin{subfigure}[b]{0.48\columnwidth}
        \centering
        \includesvg[width=\textwidth]{Annotator2Change}
        \caption{Annotator 2}
        \label{subfig:ann2_change}
    \end{subfigure}
    \hfill
    \begin{subfigure}[b]{0.48\columnwidth}
        \centering
        \includesvg[width=\textwidth]{Annotator3Change}
        \caption{Annotator 3}
        \label{subfig:ann3_change}
    \end{subfigure}

    \caption{The change in annotators' scores between the first and second annotation. No difference: initial difference between annotators < 3 points; Moderate difference: initial difference 24--26; High difference: initial difference > 50.}
    \label{fig:ann_change}
\end{figure}

\subsection{Results}
The inter-annotator agreements in Pearson's correlations for the re-annotation subset are shown in Table~\ref{tab:reannotation}.
The initial correlations were low, ranging from 0.089 (A1 vs. A2) to 0.237 (A2 vs. A3). For both annotators who re-annotated the texts, the correlation between their first and second annotations (Annotation vs. Re-annotation) was in the moderate range. Notably, the re-annotation of the same subset produced a much higher agreement between the two annotators (Pearson's \emph{r} 0.678), exceeding even the correlations between each annotator's own initial and re-annotation scores. 

As shown in Figure~\ref{fig:ann_change}, the degree of difference between the original annotations appears to influence how much each annotator's original annotation deviated from their initial one.
For both annotators, the larger the initial disagreement, the greater the change in their re-annotations---as expected. Even the texts that initially had little to no difference between annotators (the control set) exhibited some changes, although most of these were within a reasonable range of up to 25 points. Only two texts in this category received notably higher scores, both from A2.

\begin{table}[!t]
    \centering
    \small
    \begin{tabular}{lrr}
        \toprule
        & \bf Pearson's \emph{r} & \textit{p}-value \\
        \midrule
         \bf Initial annotation \\
         \cmidrule(lrr){1-3}
         A1 vs. A2 & -0.089 & 0.160\\
         A1 vs. A3 & -0.237 & < 0.001\\
         A2 vs. A3 & 0.148 & 0.019\\
         \midrule
         \bf Re-annotation \\
         \cmidrule(lrr){1-3}
         A2 vs. A3 & 0.678 & < 0.001\\
         \midrule
         \multicolumn{3}{l}{\bf Annotation vs. Re-annotation} \\
         \cmidrule(lrr){1-3}
         A2 & 0.579 & < 0.001\\
         A3 & 0.419 & < 0.001\\
         \bottomrule
    \end{tabular}
    \caption{Pearson correlations for the re-annotation subset of 250 texts.}
    \label{tab:reannotation}
\end{table}

\subsection{Qualitative Insights}
As one annotator noted in their post-annotation interview, the difference between their first and second annotations, as discussed in the previous subsection, might have been influenced by context. When annotating several subjective texts in a row, the annotator felt they might have subconsciously started to look for signs of objectivity and, therefore, rated subsequent texts as more objective than they would have if those texts had been preceded by objective ones. This suggests that context---in this case, the other texts in the batch---influences how humans perceive and assess subjectivity. 

The potential impact of context was also mentioned in another sense, concerning the annotators' physical location.
Because LimeSurvey is an online platform, annotators could complete their tasks from any location. Each batch of 250 texts was annotated over the course of roughly one week, allowing participants to choose convenient settings for their work. As one annotator noted, working from home might have led them to produce slightly different ratings than they would have in a laboratory environment.

While both annotators said that they recognized the content of some of the texts (for instance, a news story about a housefire, as well as a forum post about video games that was heavy on terminology), they did not remember how they had placed these texts on the scale.
This means they neither attempted to replicate nor change their initial annotations. 
Additionally, one annotator described their annotation process as follows: they initially analyzed the text by looking at the adjectives, after which they analyzed the objective of the text.
If the objective was to sell a product, they stated that they felt the need to mark the text as more subjective, even if the typically subjective adjectives were not present in the text.

When asked what might have caused low or partial assuredness, the annotators mentioned text length as a main cause of uncertainty. However, when looking at the annotations themselves, this pattern between text length and certainty did not emerge. It is possible that these texts fell into an in-between category, where the annotator was not completely certain, but they were more than halfway certain, showcasing the lack of nuance of our chosen 3-point Likert scale. As another aspect impacting certainty, one annotator brought up the philosophical question of what constitutes subjectivity, citing the topic of human rights: for instance, is the statement ``all humans have a right to clean drinking water'' objective or subjective?

\section{Fourth Annotator}
Following the initial analysis, which revealed relatively high disagreements between annotators, 
and a subsequent re-annotation of a subset of texts that showed higher inter-annotator agreement than the initial annotations,
a fourth annotator was recruited to the project. 
The main motivation for adding this annotator was to move toward a more reliable aggregation of subjectivity scores across several annotators, particularly in cases where the three initial annotators had shown substantial disagreement. While including additional annotators could further improve the stability of these aggregated scores, this would be prohibitively costly.
This annotator had also participated in the pilot project. During late May and early June of 2025, she was given the full dataset of 1000 texts in four batches, under the same conditions as the initial three annotators, including compensation of 1000€.

The Pearson correlations between the fourth annotator (A4) and the initial annotators ranged between a narrow interval, from 0.627 (A1 vs. A4) and 0.675 (A2 vs. A4), indicating similar consistency across all three comparisons. The \textit{p}-values for all correlations were below 0.001.

The correlations with the secondary annotations on the re-annotation subset were lower---0.420 with A2 and 0.403 with A3 (\textit{p}-values both below 0.001)---although these values were still much higher than the correlations observed among the initial annotations within the same subset. In addition, A4 reported the lowest level of assuredness overall, marking 392 texts as ``not at all certain`` and 23 as ``partially certain``.

Overall, adding the fourth annotator provided useful additional data for aggregating subjectivity scores and improving dataset reliability.
The relatively high correlations between A4’s annotations and those of the initial annotators suggest that their evaluations align well with the existing data, while their lower assuredness may provide interesting material for future qualitative research on how uncertainty influences the perception of subjectivity.

\section{LLM Annotation}

After finishing the human annotations and discovering the intricacies within them, questions of automatic subjectivity analysis arose: if this is a task where humans provide varied scores, how would large language models (LLMs) handle it?

\subsection{Initial trial}
To test whether LLMs can produce usable subjectivity annotations, we first prompted several models with eight texts representing different levels of subjectivity. Only texts with low disagreement between annotators were selected. All tested LLMs produced scores that were fairly similar to those of human annotators, even without providing example texts, differences between LLM and human scores ranged from 1 to 18. Based on these results, the outputs were considered valid, indicating that larger-scale annotation with LLMs was feasible.

During the initial LLM trial phase, the prompt asked the model not only to assign a subjectivity score but also to provide an open-ended explanation for it. This was intended to help reduce variability in the outputs and to offer additional information for prompt development if needed. The explanation was required to be in English; otherwise, it tended to fluctuate between English (the language of the prompt) and Estonian (the language of the analyzed text).

\subsection{Annotation with GPT-5}
The full set of 1,000 texts was annotated using GPT-5 \citep{GPT} with default hyperparameters.
The instruction to generate explanations was retained in the final prompt, even though these explanations were not used in the current analysis. They were kept for potential future research, as such rationales may help identify aspects that GPT-5 interprets differently from humans when analyzing subjectivity.
The prompt used is shown in Figure \ref{prompt}.
\begin{figure}
    \centering
    \begin{mdframed}
    \textrm{
    Assign the subjectivity score for the given text. A subjective text expresses the thoughts and opinions of the author, while an objective text reflects reality and expresses facts. The score for subjectivity is a value between 0 and 100, where 0 is a very objective text and 100 is a very subjective text. Texts with a score anywhere between 0 and 100 contain some amount of both subjectivity and objectivity. A subjective text does not have to be emotional, so ignore lack of emotionality as a reason to reduce the subjectivity score. Additionally provide a free format explanation in English as to why this specific score was given and not a higher or lower one.}
    \end{mdframed}
    \caption{Instructions used to prompt GPT-5.}
    \label{prompt}
\end{figure}

While formatting was not included in the prompt itself, it was included in the API task request as a JSON schema, where there were two parameters: score, or the subjectivity evaluation, and reason, or the freeform rationale.

To see the impact of randomness, the dataset was annotated three separate times using batching, with each time the set of 1000 texts being a different batch. The scores provided by GPT-5 did vary, with subjectivity scores being different between batches by up to 40 points. However, the Pearson correlation coefficients between the batches were between 0.978 and 0.979, demonstrating considerable consistency. Thus, we averaged the scores over three runs to get the final LLM predictions.

Although the three GPT-5 runs showed high overall consistency, some variation between batches was still observed. A total of 146 texts differed by more than 10 points, and 16 by more than 25 points (the threshold used to indicate a moderate difference between human annotators), with the largest difference reaching 40.
Most texts with larger differences were narrative in nature, with the text showing the greatest difference (40) being a short folkloric legend and the third most varied text (34-point difference) consisting of short plot summaries of two movies. This pattern suggests that GPT-5 may be less consistent in evaluating the subjectivity of narrative texts.

\subsection{Comparison to human annotations}
When comparing the LLM scores to human annotators, the correlations were similar to those between the human annotators, varying between 0.601 and 0.801, with \textit{p}-values for all annotators being below 0.001. 
However, when looking at the differences qualitatively, 
additional patterns started to emerge. The most significant difference between humans and LLM was the range of scores given. While humans used the full range of 0 to 100, GPT-5 never ended up giving a score higher than 98.

Looking at the texts where the differences between the GPT-5 and human scores were the highest, two aspects stand out---quotes and colloquial language use. Quotes, a method to directly and accurately refer to sayings and statements made by third parties, are commonly used in news stories. Many of the texts rated highly subjective by GPT-5 and low, as objective, by human annotators were news stories containing quotations. It is probable that human annotators were able to understand the intention of the quotes, to refer to a statement by the person discussed in the story, and therefore ignore any potential subjectivity within them. Meanwhile, GPT-5 considered the content of these quotes as a part of the text, causing the subjective score to rise. This was most apparent in shorter news stories that primarily consisted of quotations, including the text with the highest difference between the mean human score and the mean GPT-5 score (a difference of 70.3 points).

Colloquial language use, specifically the language used in forums and blogs that prominently includes aspects such as slang, emoticons, and generally simpler syntax, tended to heavily increase human subjectivity scores. However, GPT-5 tended to ignore this change in tone and keep the scores lower, more objective. This meant that while colloquial language tended to influence humans, GPT-5 focused more on the content of these texts.

While these correlations prove GPT-5 can produce plausible subjectivity scores for web texts, further analysis is needed to ensure their validity. 
Manual comparison of the scores with the corresponding texts allows differences between human and GPT-5 annotations to be examined in detail.
If consistent patterns of divergence emerge, they may show that GPT-5 cannot yet serve as a reliable substitute for human subjectivity annotation.

\section{Discussion}

The results of both the pilot and the full annotation study showed that human annotators were able to use this scale consistently, although the inter-annotator correlations were moderate. The Estonian subjectivity dataset created through this process, therefore, contains valid and useful data and also shows that there is still more to analyze and potentially improve in the future. In particular, while several aspects were identified that were not found to be factors to low correlations between annotators, such as text length and categories, the potential reasons for differences in annotators' scores require further qualitative analysis.

\subsection{Inter-annotator agreement}
The correlations between annotators were much higher in the pilot study than in the final dataset of 1000 texts, which only indicated a moderate correlation (from 0.525 to 0.675). Although all of the aspects that cause this reduction in correlation remain unclear, some do stand out as possible causes. 

The first of these aspects is the texts themselves: while there were only 60 texts in the pilot project, the 1000 texts in the full dataset provide a much larger variety of topics, writing styles, and text types. It is possible that a different set of texts would have led to different interannotator correlations. Additionally, it is likely that some texts are inherently more difficult to annotate and analyze. These texts would always get differing scores from the annotators, leading to low correlations. However, these texts should not be simply discarded from the dataset, as these disagreements could provide for interesting research data on what aspects of text make a document's subjectivity and objectivity difficult to understand.

The second aspect that may have reduced interannotator agreement is who the annotators were: neither of the annotators in the pilot project were initially included in the annotation of the 1000 texts; therefore, these comparisons are not just made with a different set of texts but also with different people. With more annotators, at least one pair of annotators with a similarly high correlation may emerge. This can partially be seen with the addition of the fourth annotator, who correlates the most with all of the previous annotators.

\subsection{Context effects}

One possible explanation for the observed differences between annotations is the influence of context, that is, the previously seen texts. During the re-annotation phase, one annotator noted that after scoring several highly subjective texts in a row, they felt more sensitive to objectivity and tended to assign lower subjectivity scores to the following texts than they might have if those texts had been preceded by more objective ones. This suggests that human annotators are influenced by the sequence in which texts are presented and that the score of any given text may depend on what has been seen immediately before it.

Because the texts were presented in a random order for each annotator, none of the annotators shared the same contextual sequence, and this variation in context may partially explain the differences in their scores. Similar sequence or order effects have been reported in other subjective annotation tasks, such as the rating of emotional intensity \citep{mathur2017sequence}, sentiment of product reviews \citep{toker2024intelligent}, and hate speech and offensive language \citep{beck2024order}. It is therefore plausible that similar context effects also occur in subjectivity annotation, although the precise mechanisms behind them require further investigation.

\subsection{Human vs LLM subjectivity}

For humans, subjectivity analysis is, ironically, a subjective task, where one can focus on either the subjectivity of the language used or the content of the itself, or look at both simultaneously---the choice is up to the analyzer. Additionally, it is not always clear what is considered objective truth. For instance, one annotator pointed out during the interview that they were unsure whether common beliefs, specifically the topic of human rights, should be regarded as objective. Fortunately, despite these aspects, people will generally be on the same page when analyzing the subjectivity of a text.

In addition to human annotations, we experimented with using an LLM, GPT-5, to automate the annotation process. To verify consistency, GPT-5 was prompted three times, with the correlation between the batches averaging 0.98, showing very high consistency. One aspect that lowered this consistency was narratives, such as descriptions of what a person did during their day, which appeared more difficult for the model to evaluate reliably.

When comparing human annotations with GPT-5-generated outputs, additional patterns emerged. 
First, GPT-5 rated texts with quotations, such as news stories, consistently higher than human annotators did. Secondly, human annotators rated texts written in a very informal tone---especially those using language common in forums and blog posts---as highly subjective, whereas GPT-5 focused more on the content of those texts, giving less weight to tone. Similar systematic differences between human and LLM annotations have also been reported in other subjective labeling tasks, such as emotion labeling \citep{niu2025rethinking,bojic2025comparing,greschner2025fearful} and safety ratings \citep{movva2024annotation}.

Because of these differences, GPT-5 was not able to fully reproduce human-like annotations. However, the correlations between human annotators and the GPT-5 outputs ranged between 0.593 and 0.798, which is comparable to the correlations observed between human annotators themselves. This indicates that GPT-5 scores are plausible subjectivity values that can be used in applications where automation and speed are important.
Still, since GPT-5 and human annotators rely on partly different cues—content versus tone and language use—LLMs cannot yet serve as a replacement for human subjectivity annotation, particularly in linguistic or interpretive research where human-like understanding is essential.

\section{Conclusion}
This study introduced a new approach to subjectivity annotation by creating a dataset of 1,000 Estonian texts rated on a continuous numeric scale from 0 (fully objective) to 100 (fully subjective). The dataset combines human annotations with GPT-5-generated scores, providing a valuable resource for both linguistic research on subjectivity and the development of automatic subjectivity detection systems.
The results show that human annotators can use the numeric scale consistently, although their judgments may be influenced by contextual factors such as the sequence of texts. The comparison between human and GPT-5 annotations further demonstrates that while the model produces plausible subjectivity scores and correlates moderately with human ratings, it also relies on somewhat different cues. These differences highlight that LLMs may complement but not yet replace human annotators in subjective tasks.
Future work should expand the dataset with additional human annotations to examine the effects of context more systematically and to establish more robust aggregated subjectivity scores. Further experimentation with prompt design, other LLMs, and example-based prompting may also help to better understand and reduce the observed human–LLM discrepancies.

\section{Acknowledgments}
We would like to acknowledge the work of the annotators and thank them for their contributions to the dataset. This work was partially supported by the National Program for Estonian Language Technology (project EKTB104) funded by the Estonian Ministry of Education and Research and by the Estonian Research Council Grant PSG721.

\nocite{*}
\section{Bibliographical References}\label{sec:reference}
\bibliographystyle{lrec2026-natbib}
\bibliography{lrec2026-example}

\clearpage
\appendix
\renewcommand{\thesection}{Appendix \Alph{section}}
\section{Instructions given to annotators translated to English}

Hello!

The goal of the project is to assess the subjectivity of full texts on a scale. Subjectivity means that a text expresses the author’s opinions and attitudes. Its opposite is objectivity, where a text expresses facts corresponding to the real world.

As part of the project, we ask you to annotate 1,000 texts in four batches, each batch containing 250 texts. The batch to be annotated is the first\footnote{The numeration was changed according to batch.} one.

Please annotate subjectivity on a scale from one to one hundred. On this scale, 0 is a completely objective text and 100 a completely subjective text. Texts that contain both subjective and objective elements are represented by intermediate numbers.

This is a subjective task, meaning that no text has a single correct numerical subjectivity rating, since each person perceives it differently. The aim of the annotation is to obtain several different assessments for each text, and we will use the averages of these assessments to build a machine-learning model that evaluates subjectivity.

It is possible to add comments to each text. Please write there any thoughts you have regarding the annotation itself, especially any issues or problems. There is no need to specify which aspects of the text influenced your rating. If you have questions that you would like answered immediately, please write to [EMAIL].

On the next page, please write a name, either your own or a pseudonym, so it would be possible to match annotators across different batches. In the final, formatted dataset this will not be published; instead, it will be replaced with a random, unrelated identifier.

\section{Example texts translated to English}
Following are three illustrative sample texts — one strongly objective, one strongly subjective, and one that is moderate.

\subsection{Example of an objective text}
Long-term supported employment service
The long-term supported employment service is intended for unemployed people who have been identified as having no or partial work ability and who need extensive preparation before they are ready to move to the open labor market; as a result, they may remain in the supported employment service for a long period. The goal of the LTSE service is to offer people with reduced work ability an opportunity to perform work suited to their capacity in a protected and adapted environment, as well as, if necessary, support in transitioning to labor market services or employment in the open labor market. Working under protected conditions allows people who need greater support to acquire work-related skills in an environment that takes their needs into account and at a pace appropriate for them, with guidance and assistance provided to the necessary extent.

\subsection{Example of a subjective text}
Sunday, 16 December 2007
About making gifts

I went yesterday to my nephew’s bday. He turned 2 years old. But naturally I couldn’t go without a present. Therefore I visited a toyshop. If last year the search for a present for a one-year-old was a crazy trouble, then now it went already easier. Last year I also over-did it and got confirmation that the old folks didn’t say their sayings for nothing. You really shouldn’t buy a pig in a poke. It can turn out a calf or a kitten.

Last year’s present, a colorful soft worm, which was supposed to also happily sing, turned out actually to be a quietly buzzing sausage-like creature. There was no joy from it for the kid nor for me. And it cost quite nicely too.

This time I bought a simple dump-truck. The boy is already starting to show interest for more manly themes and why not. As an addition to the collection of four-wheeled movers, because from New York I brought him a police car. Of course he was more interested in the snoring and dancing puppies that his sister gifted, but whatever. At least it will stay in active use.

About the bday… Well, it was probably the last one which the boy himself still doesn’t understand a thing of. The birthday kid did his small-kid things and felt good most of the time. And we ate and ate and ate :).

\subsection{Example of a moderate text}
The sauna-bus is relaxing and very comfortable. Thanks to its mobility you can enjoy sauna pleasures wherever you want! The sauna-bus’s sauna bench fits 5 people, the steam room also has a spacious front room. You don’t have to heat the sauna-bus yourself, our specialists take care of that and they do it completely unnoticed, as the sauna stove opens from outside. In the sauna-bus there is a shower, WC, sink, fridge, cupboards and a stereo system. Alongside the sauna it is possible to also get a party room which fits 50 people. With a proper sound system and a big screen for video-disco, karaoke, sports broadcasts on TV (3x2.5m screen) i.e. for watching picture galleries. On the spot it is also possible to make a winter grilling evening. Located on Sossi hill in Tallinn and if you have sauna’d enough and want to go somewhere further then within a couple hundred meters there are plenty of places to go to.
\onecolumn
\section{Table of Data Included in Final Dataset}

\begin{table*}[h]
    \small
    \centering
    \begin{tabularx}{\textwidth}{llX}
        \toprule
         \bf Column Name & \bf Description & \bf Information \\
        \midrule
         ID & Unique identifier & 1000 unique strings \\
        Text & The full text & 1000 unique strings \\
        Category & The text category or genre & 11 unique strings \\
        Mean Human Score & \makecell[l]{Average subjectivity score based on all\\four annotators for full 1000 text dataset\\(additional annotation subset excluded)} & Range 0---100, float \\
        Annotator 1 & \makecell[l]{The subjectivity scores given by\\A1 for full 1000 text dataset} & Range 0---100, integer \\
        Annotator 2 & \makecell[l]{The subjectivity scores given by\\A2 for full 1000 text dataset} & Range 0---100, integer \\
        Annotator 3 & \makecell[l]{The subjectivity scores given by\\A3 for full 1000 text dataset} & Range 0---100, integer \\
        Annotator 4 & \makecell[l]{The subjectivity scores given by\\A4 for full 1000 text dataset} & Range 0---100, integer \\
        Annotator 1 Certainty & \makecell[l]{A1 confidence in their score for\\each text in full 1000 text dataset} & 3-point Likert scale \\
        Annotator 2 Certainty & \makecell[l]{A2 confidence in their score for\\each text in full 1000 text dataset} & 3-point Likert scale \\
        Annotator 3 Certainty & \makecell[l]{A3 confidence in their score for\\each text in full 1000 text dataset} & 3-point Likert scale \\
        Annotator 4 Certainty & \makecell[l]{A4 confidence in their score for\\each text in full 1000 text dataset} & 3-point Likert scale \\
        Annotator 2 Addition & \makecell[l]{The subjectivity scores given by A2\\for the 250 text re-annotation subset} & Range 0---100, integer \\
        Annotator 3 Addition & \makecell[l]{The subjectivity scores given by A3\\for the 250 text re-annotation subset} & Range 0---100, integer \\
        Annotator 2 Addition Certainty & \makecell[l]{A2 confidence in their scores for\\the 250 text re-annotation subset} & 3-point Likert scale \\
        Annotator 3 Addition Certainty & \makecell[l]{A3 confidence in their scores for\\the 250 text re-annotation subset} & 3-point Likert scale \\
        Mean GPT Score & \makecell[l]{Average subjectivity score based on\\all three GPT prompting batches\\for full 1000 text dataset} & Range 0---100, float \\
        GPT Score 1 & \makecell[l]{Batch 1 subjectivity scores given by\\ GPT-5 for full 1000 text dataset} & Range 0---100, integer \\
        GPT Score 2 & \makecell[l]{Batch 2 subjectivity scores given by\\ GPT-5 for full 1000 text dataset} & Range 0---100, integer \\
        GPT Score 3 & \makecell[l]{Batch 3 subjectivity scores given by\\ GPT-5 for full 1000 text dataset} & Range 0---100, integer \\
        GPT Explanation 1 & \makecell[l]{Batch 1 rationales for subjectivity scores\\given by GPT-5 for full 1000 text dataset} & 1000 unique strings \\
        GPT Explanation 2 & \makecell[l]{Batch 2 rationales for subjectivity scores\\given by GPT-5 for full 1000 text dataset} & 1000 unique strings \\
        GPT Explanation 3 & \makecell[l]{Batch 3 rationales for subjectivity scores\\given by GPT-5 for full 1000 text dataset} & 1000 unique strings \\
        Number of Characters & Number of characters in full text & Integer \\
        Number of Words & Number of words in full text & Integer \\
        Number of Sentences & Number of sentences in full text & Integer \\
        Batch & \makecell[l]{Indicator for the batch in which\\the text was sent to the annotators} & 4 unique strings \\
        Original Metadata & Text metadata from Estonian National Corpus& 1000 unique strings \\
        \bottomrule
    \end{tabularx}
    \label{tab:datasetcolumns}
\end{table*}

\end{document}